\pdfoutput=1
\documentclass[sigconf,nonacm]{acmart} 
\renewcommand\footnotetextcopyrightpermission[1]{} 

\acmConference[ ]{ } 
\acmBooktitle{ } 
\usepackage{graphicx} 
\graphicspath{{img/}}
\usepackage{booktabs} 
\usepackage{array} 
\usepackage{caption} 
\usepackage[title]{appendix}
\usepackage{algpseudocode} 
\usepackage{amsmath,amsfonts}
\usepackage{multirow}
\usepackage{tablefootnote}
\usepackage{wrapfig}
\usepackage{arydshln}
\usepackage{url}
\usepackage{subcaption}
\usepackage{float}
\begin{document}

\title{Explainable and Interpretable Forecasts on Non-Smooth Multivariate Time Series for Responsible Gameplay}

\author{Hussain Jagirdar}
\email{hussain.jagirdar@games24x7.com}
\affiliation{%
  \institution{Games24x7}
  \city{Bengaluru}
  \country{India}
  }
\author{Rukma Talwadker}
\email{rukma.talwadker@games24x7.com}
\affiliation{%
  \institution{Games24x7}
  \city{Bengaluru}
  \country{India}
  }
\author{Aditya Pareek}
\email{aditya.pareek@games24x7.com}
\affiliation{%
  \institution{Games24x7}
  \city{Bengaluru}
  \country{India}
  }   
\author{Pulkit Agrawal}
\email{pulkit.agrawal@games24x7.com}
\affiliation{%
  \institution{Games24x7}
  \city{Bengaluru}
  \country{India}
  }  
\author{Tridib Mukherjee}
\email{tridibm@gmail.com}
\affiliation{%
  \institution{Games24x7}
  \city{Bengaluru}
  \country{India}
  }

\begin{abstract}
Multi-variate Time Series (MTS) forecasting has made large strides (with very negligible errors) through recent advancements in neural networks, e.g., Transformers. However, in critical situations like predicting gaming overindulgence that affects one's mental well-being; an accurate forecast without a contributing evidence (explanation) is irrelevant. Hence, it becomes important that the forecasts are \textit{Interpretable} - intermediate representation of the forecasted trajectory is comprehensible; as well as \textit{Explainable} - attentive input features and events are accessible for a personalized and timely intervention of players at risk. While the contributing state of the art research on interpretability primarily focuses on temporally-smooth single-process driven time series data, our online multi-player gameplay data demonstrates intractable temporal randomness due to intrinsic orthogonality between player's game outcome and their intent to engage further. We introduce a novel deep \textit{Actionable Forecasting Network (AFN)}, which addresses the inter-dependent challenges associated with three exclusive objectives - 1) forecasting accuracy; 2) smooth comprehensible trajectory  and 3) explanations via multi-dimensional input features while tackling the challenges introduced by our non-smooth temporal data, together in one single solution. AFN establishes a \textit{new benchmark} via: (i) achieving 25\% improvement on the MSE of the forecasts on player data in comparison to the SOM-VAE based SOTA networks; (ii) attributing unfavourable progression of a player's time series to a specific future time step(s), with the premise of eliminating near-future overindulgent player volume by over 18\% with player specific actionable inputs feature(s) and (iii) proactively detecting over 23\% ($\sim$100\% jump from SOTA) of the to-be overindulgent, players on an average, 4 weeks in advance.
\end{abstract}
\maketitle

\section{Introduction}
In the domain of online skill gaming such as Rummy~\cite{Rummy_Wikipedia}, regular analysis of player's gaming behavior is crucial to detect and address  overindulgence - consistent and sustained engagement at an extreme level ~\cite{scarcegan,gambling1,gambling2}. Overindulgence can lead to extreme outcomes for a player arising out of monetary losses and mental exhaustion. Sustained healthy engagement ensures better long term rewards, both for the player as well as the business. Although occurrences of extreme risky events are  scarce (with 0.05
- 0.1\% probability) ~\cite{sigir}, it is important to identify them in-time to ensure and encourage responsible game play with proper intervention. This necessitates not only in-time monitoring but a proactive and timely forecasting of players' multi-variate time series data. Further, to action on a prediction of overindulgence, for a player who has been playing moderately (safely) in the present times, is difficult to comprehend unless the trajectory of the prediction is human interpretable and can be explained via shifts in the player's one or more game play features. 

\subsection{Background and Challenges:}
\noindent\textbf{A. MTS Forecasting on Multi-Source Highly Random Data}

MTS forecasting has made large strides through recent advancements in neural networks, e.g., Transformers, both, in multi-time step ~\cite{crossformer,scaleformer,pyraformer} as well as multi horizon forecasts ~\cite{Temporal_Fusion_Transformers}. Some of the recent and directly applicable research on interpretable representation learning on time series data ~\cite{som-vae, DPSOM} has furthered the importance of interpretability at par with forecasting for proactive intervention, especially in the health domains. However, much of these networks assume temporally smooth transitions in the data properties and primarily demonstrated results on synthetic data where temporal smoothness was induced via linear interpolation methods. We contend that these existing SOTA networks are insufficient towards our objective of actionable forecasts, mainly due to the following distinct characteristics observed in our player data:\\
(i) \textbf{Highly Random and Temporally Non-Linear Progression:} Though gaming patterns are quite outcome based (win/loss), the psychological imprints on the player from their previous game play could generate very diverse game behaviours in immediate next play (\cite{cognitionnet,ARGO}) - player with previous losses could show assured and confident game play style while somebody with high winnings may actually show unnecessary aggression, and vice versa. Hence the factors (co-variates) affecting the future game play decisions are not completely observed, measurable or generalizable. Due to which the gameplay data arising out player's time-varying psychological imprints mostly posse high irregularity. This presents itself as our primary and a unique challenge. Section \ref{sec:random_evaluation} comprehends data randomness quantitatively. 

(ii) \textbf{Multi-process Nature of Data Generation:} Much of the SOTA forecasting networks have been evaluated on a data (WTH \cite{wth}, ECL \cite{ecl}, ETTh, ETTm \cite{ett} etc.) generated from a \textit{single-process} - e.g electricity demand of a particular city, temperature of a specific locality, over a time period. Our data comes from multi-million active players on a daily basis - synonymous to many processes and its generalization is also a challenge.


\noindent\textbf{B. Interpretability} As much as the accuracy over the forecasts matter, it's important for the forecasts to be \textbf{interpretable}, for e.g visualization of the entire trajectory in a lower dimensional space (2-D) where the human cognition works the best. For example, via using a high dimensional eICU data ~\cite{eicu,eicu-crd-2.0} which contains various measurements related patient's physiology, a model predicts fatality in the next few weeks; a comprehensible trajectory should help explain when in time and how the model made that prediction. Interpretable forecasts are a necessity in many domains, especially involving rare/outlier occurrences like fraudulent transactions \cite{kaggle}, network intrusion \cite{icdm,intrusion2}, digital or social media addiction \cite{social,netflix}, gambling \cite{gambling1,gambling2}, addictive e-commerce shopping \cite{shopaholic} etc.

Most of these time series models are based on a encoder-decoder architecture, that maps the high dimensional input data into a latent space primarily making an i.i.d. assumption about the temporal data, ignoring the rich temporal structure available in the adjacent data time steps. Given that human cognition is not inherently optimized for efficient functioning in high-dimensional spaces SOM-VAE~\cite{som-vae} and T-DPSOM ~\cite{DPSOM} based models propose to further reduce this intermediate latent representation space into a two-dimensional grid enforcing Self Organised Map (SOM) ~\cite{som} like topological neighbourhood. Interpretability is then facilitated by encouraging a higher probabilistic likelihood that temporally adjacent data points belong to the same or immediately next SOM centroid, assuming linear progression in data properties over time. This is done via a probabilistic transition model in \cite{som-vae} and a LSTM based model in \cite{DPSOM}. However, irregularity in temporally adjacent data points arising out of random process, yields following challenges:
(i) \textbf{Temporally Adjacent data points appear Topologically Distant:} data randomness contradicts the temporal smoothness assumption made by the SOTA networks. When trained using these networks our temporally adjacent data points appear far away noticeable number of times. This obstructs interpretability as temporal trajectories appear to move randomly. (ii) \textbf{Lack of Evidence on Sudden Long Jumps in the Representation Space:} Assuming this pattern of randomness is generally observed across multiple players, we could enforce SOM embedding space using the prior art. However, lack of \textit{quantifiable} evidence or justification for a distant jump, could render the interpretability difficult to accept and initial further proactive action.

\noindent\textbf{C. Explainability} In addition to interpretability, its important to a) identify the significant feature(s) that are dominant for a player at a particular time-step and b) time steps in the near future which are attentive towards players future overindulgence. Together, both these would aid in personalizing the proactive-actions that aligns with the player's well-being. SHAP~\cite{SHAP} (SHapley Additive exPlanations) and LIME (Local Interpretable Model-agnostic Explanations)~\cite{LIME} are popular techniques towards explainable AI (XAI). However, conventional use of these model-agnostic post-hoc methods has following limitations: (i) \textbf{Negligence of Temporal Ordering of Input Features in Time Series:} This can lead to sub-optimal explanation quality, as temporal dependencies are often significant in time series data, (ii)  \textbf{Shifting focus on attentive features curbs accurate treatment:} shifting attention on features over the time period of intervention would confuse treatment plans; e.g. patient being treated with a different drug every day due to shift in  attention weights between various physiology parameters. 

\subsection{AFN Intuition and Contributions:}
Towards building a robust actionable forecasting network while addressing the inter-dependent challenges associated with our three, equally important but exclusive objectives - forecasting, interpretability and temporally consistent explainability we make certain design decisions. These decision focus on \textit{learning hidden} aspects related to random temporal data drifts via leveraging some existing research in the player psychology space. 

\textbf{Estimating and Leveraging Temporal Randomness in data:} We proceed with a reasoning that, shifts in the game-play data of the players, occur due to a change in their psychological imprints. We hypothesise that the \textit{transition} in these imprints are Markovian in nature. We can then associate psychological imprints to \textit{latent states} in a Markov chain. AFN models these stochastic relationships between latent states via a \textbf{Deep Markov Model (DMM)}, a deep neural network. Thereby, given a latent state for an immediate past time step, DMM predicts the present one. This outcome is further mapped to a discrete quantity (scalar), via another Conditional Network. These \textit{conditions} helps in identifying the parameters of the present data distribution (given the immediate past one). We could extend this to k-step Markov as well. Using this, we attempt to learn and predict irregularities in temporal data per player via a  \textbf{Transition Module (TM)}.

\textbf{Extending SOM based Interpretability for Temporal Irregularities:} Now, since the network is aware of the psychology of the player in the form of the condition predicted by the TM, we target to enrich the latent space of the downstream VAE~\cite{vae} based encoder-decoder network as in~\cite{som-vae, DPSOM} with this condition parameter. This helps in \textit{competitive} mapping of variably scaled, non-smooth feature values (arising out of different conditions and hence different distributions) to a common intermediate (latent) space of the VAE over which similar SOM based topology could be extended. Word `competitive' here refers to the objective which strives to balance between topological smoothness- for similar data points via SOM property vs. retention of interpretability - by avoiding long jumps when data points are coming from different distributions and hence dissimilar. This is done via appropriate learning objectives in AFN. Since the conditions are learnt and inferred at runtime, we refer to it as ConVAE as opposed to the traditional cVAE. 

\textbf{Putting it all together:} We propose \textbf{Intelligent Forecasting Module (IFM)} that models sequential time series data, by training a \textbf{LSTM}~\cite{lstm} on the latent encodings of temporal data, produced by ConVAE-SOM. Training on these encodings allows the model to retain interpretability as these latent encodings are mapped to their respective SOM clusters. \textbf{Attention Layer (AL)} is added on top of LSTM to focus on the relevant time steps. ConVAE-SOM encourages a higher probabilistic likelihood for latent representations belonging to temporally adjacent data points to the same or immediately next SOM centroid. To accommodate temporal non-smoothness arising due to data randomness, we provision a `\textit{leeway}' for a shift in the TM condition parameter. This is done via a \textbf{Damping Factor Network (DF)} which is trained to self-learn a damping factor to help the network balance smoothness over the prediction loss. AFN  re-establishes \textbf{SHAP} ~\cite{SHAP} over the SOM clusters by blending attentive time steps from the AL to ensure consistent feature focus for proactive intervention. We summarize the main \textbf{contributions} of AFN as follows.

\noindent \textbf{1. Transition Module (TM)}: A Deep Markov Model and a Conditional Network which attempts to encapsulate the scale varying features and their periodic/seasonal variations into a transition function to provide relevant conditions to ConVAE-SOM. This alleviates the challenge of predicting \textit{randomness} arising out of \textit{hidden} and not easily measurable factors.

\noindent  \textbf{2. ConVAE-SOM}: Re-parameterization of the representations (latent space) via a condition parameter which is learnt by the TM.

\noindent  \textbf{3. Intelligent Forecasting Module (IFM)}: Joint optimization of the smoothness loss to discourage long jumps towards preserving interpretability and a damping factor which provides a “leeway” to perform a non-smooth, far away jumps for forecasting accuracy and to further quantify \textit{non-obvious} interpretability .
Our code base and the relevant datasets are available at ~\cite{AFNgit}.

\section{Related Work}
\textbf{Multivariate Time Series Forecasting:} MTS forecasting models are roughly divided into statistical (Vector auto-regressive (VAR) model~\cite{var}, Vector auto-regressive moving average (VARMA)) and neural networks. Statistical models assume a linear cross-dimension and cross-time dependency. With the development of deep learning, many neural models often empirically show better performance than statistical ones. LSTnet ~\cite{LSTnet} and MTGNN~\cite{mtfgnn} use CNN and graph neural networks respectively for cross-dimension and RNN for cross-time dependencies. RNN's, however have empirically shown to have difficulty in modeling long-term dependency. 
Recently, many Transformer\cite{transformer, sparce_binary_transformers}-based models have been proposed for MTS forecasting, showing great results. Informer~\cite{informer} proposes ProbSparse self-attention which achieves $\mathcal{O}(L \log L)$ complexity. Pyraformer~\cite{pyraformer} introduces a pyramidal attention module that summarizes features at different resolutions and models the temporal dependencies of different ranges. Following up with the improvements, Crossformer ~\cite{crossformer} further focused on the cross-dimension dependencies which further pushed the MSE baselines to an extra-ordinary level. Though quite impressive, these works do not offer any explanations over the prediction space, except ~\cite{Temporal_Fusion_Transformers} which extracts the overarching trends and patterns from the data, rather than delving into the specific causality of predictions on a case-by-case basis.
Lately, non-transformer based methods too have been proposed in recent literature that provide similar or even better performance with less resources\cite{TSMixer}, while some even question the effectiveness of transformer-based networks for MTS forecasting \cite{ndlinear}\\
\textbf{Interpretability:} In recent years, interpretability has increasingly been combined with generative modeling through the advent of generative adversarial networks (GANs)~\cite{gan,cgan} and variational autoencoders (VAEs) ~\cite{vae}. However, the representations learned by these models are often considered cryptic and do not offer the necessary interpretability~\cite{infogan}. A lot of work has been done to improve them in this regard ~\cite{beta-VAE,Hierarchical_RL}. Nonetheless, these works have focused entirely on continuous representations, while discrete ones are still being under-explored. Discrete representations are known to reduce the problem of ``posterior collapse'' \cite{VQ-VAE, VQ-VAE-2}, situations in which latents are ignored when they are paired with a powerful auto-regressive decoder typically observed in the VAE framework. \cite{som-vae,DPSOM} introduce an impressive method of achieving discrete space interpretability for high-dimensional time series, by working on SOM like clustering in a low-dimensional latent space. Major drawback being, assumption of temporal smoothness in the time series data and lacking a capability to forecast into the future.\\
\textbf{Explainability:} Explainability allows one to comprehend a particular prediction w.r.t the input features. SHAP~\cite{SHAP} assigns each feature
an importance value for a particular prediction. Traditionally, SHAP and LIME~\cite{LIME} have been used to identify feature weights for a multi-class classification or a regression based models. It still remains to be a challenge in the case of an unsupervised model, where there is no such response variable/class to hook to.\\

\section{Methodology}
Let $X = \lbrace x_1, x_2 \dots ,x_N \rbrace$ be the set of $N$ Multi-variate Time Series, each being generated by a separate process (e.g. gaming players) where each $x_i = \lbrace x_{i,1}, x_{i,2} \dots ,x_{i,T} \rbrace$ such that $x_{i,t} \in \mathbb{R}^{d}$ (where each dimension represents a player's game play feature. The objective is to forecast next $h$ time-steps $\forall x_i$ as following: $\lbrace {\hat{x}_{i,T+1}}, \hat{x}_{i,T+2} \dots$, $\hat{x}_{i,T+h} \rbrace$, given the history $\lbrace x_{i,1}, x_{i,2} \dots ,x_{i,T} \rbrace$. 

\subsection{AFN Framework}
\begin{figure*}
  \includegraphics[width=0.7\textwidth]{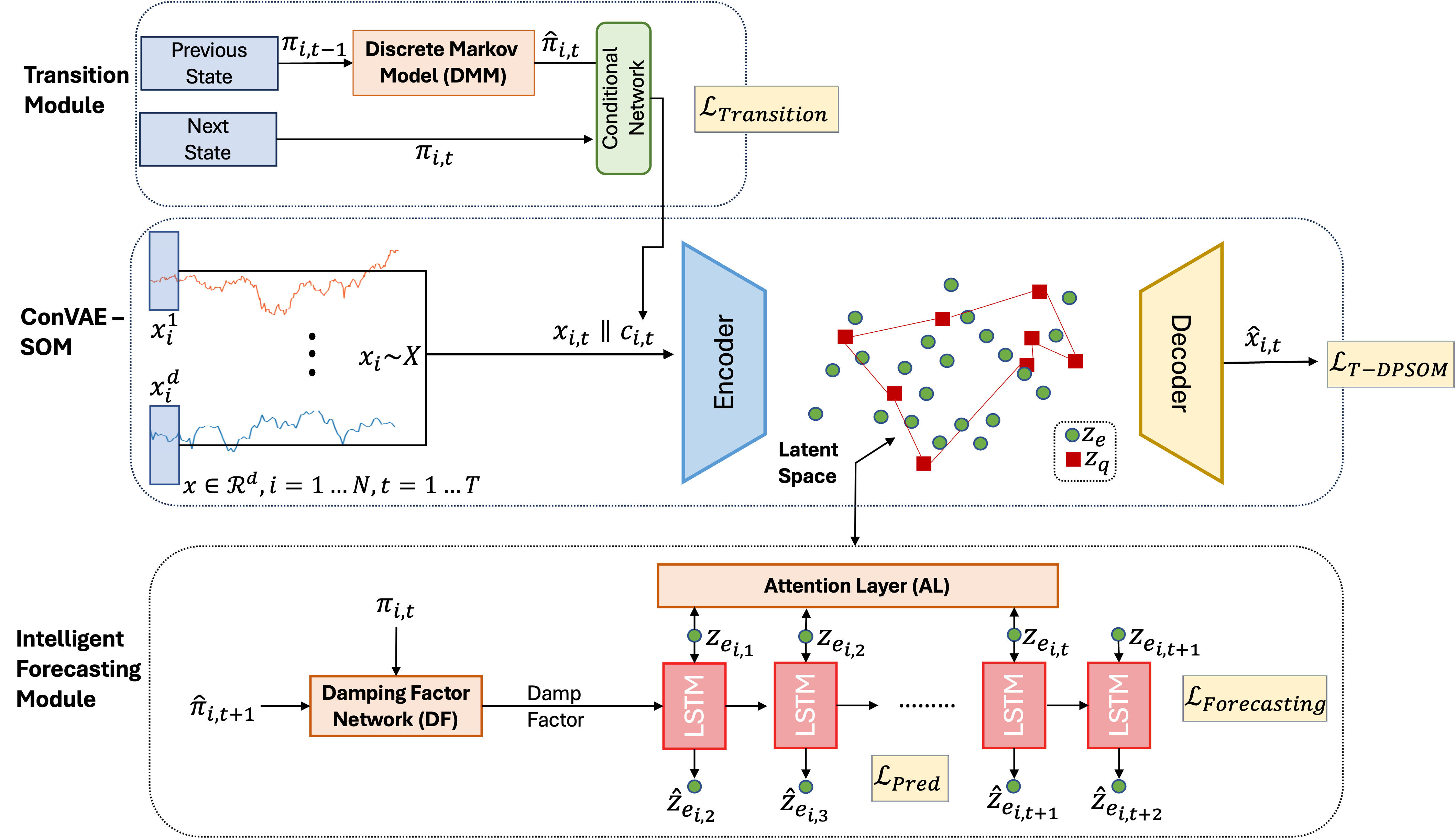}
  \caption{\textmd{AFN Architecture}}
  \label{fig:afn_architecture}
\end{figure*}

SOM-VAE\cite{som-vae} first proposed a model for learning interpretable representations over the time series data, where the authors introduce a novel concept of ``smoothness'' over temporal trajectory as an important requirement for interpretable time series models. Drawing inspiration from both \cite{som-vae} and \cite{DPSOM}, we propose AFN as a multi-variate time series forecasting model with interpretable and explainable forecasts for data with high degree of randomness (randomness quantification is done in the Section \ref{sec:random_evaluation}). Our proposed architecture is presented in Figure \ref{fig:afn_architecture}.
The major constituents of AFN are:
\begin{enumerate}
    \item Transition Module (TM)
    \item Conditional Variational Auto-Encoder with Latent SOM (ConVAE-SOM)
    \item Intelligent Forecasting Module (IFM)
\end{enumerate}

\subsubsection{Transition Module}
\label{sec:Transition_Module}
The Transition Module (TM) consisting of a Deep Markov Model (DMM) along with the Conditional Network, helps address the issue of high degree of randomness in our data. We hypothesise that the player's psychology~\cite{cognitionnet} is hidden in form of a latent state associated with each observation $x_{i,t}$, and the change in the pattern of upcoming observations is related to a change in that latent state. AFN incorporates ``Latent State Prediction'' via the DMM which is trained to predict the next state which dictates the associated pattern of the generated data. The Conditional Network works on the next state predicted by the DMM and the actual next state (available during the training phase). It discretizes the DMM's state output into a scalar ``\textit{condition}'' that influences the encodings of the ConVAE-SOM. Hence, change in these ``conditions'' leads to change in the data distribution patterns in the subsequent timestep. 

In order to describe DMM's latent state at a given time-step, we take an aggregated representation of the prior psychological imprints in a way that it encapsulates the temporal patterns. For each $x_{i,t}$, we consider the past $t-\tau$ time steps such that $\tau = C+M$, where $C$ is the  window length and $M$ is the number of windows. $C$ and $M$ are hyperparameters. We cluster each of the $M$ time windows $\lbrace {x_{i, t-\tau} \dots x_{i, t-\tau+C}} \rbrace$, $\lbrace {x_{i, t-\tau+1} \dots x_{i, t-\tau+C+1}} \rbrace$, $\dots$ $\lbrace {x_{i, t-\tau+M} \dots x_{i, t-\tau}} \rbrace$ into one of the $K$ clusters, based on a clustering technique. In AFN, we derive the clusters using CognitionNet~\cite{cognitionnet}, which are interpreted as players' dynamic psychological imprints. Alternatively, time series clustering techniques such as DTCR~\cite{DTCR}, etc. can also be used. The sequence of clusters, one for each of the $M$ time windows is summarized into a single vector $\pi_{i, t} \in \mathbb{R}^{K}$ such that each position $k \in \lbrace 1,2 \dots K\rbrace$ of $\pi_{i,t}$ represents the proportion of occurrence of cluster $k$ in the entire window of observations. Hence $\pi_{i,t}$ is a summarisation of the history of the game behaviours of player $i$ in the most recent past $\tau$ time steps. Value of $K$ -total number of unique game behaviours, is inferred by the CognitionNet.

As shown in figure \ref{fig:afn_architecture}, AFN adopts feed-forward network for both the DMM and the Conditional Network with two losses:  
\begin{enumerate}
    \item Error in DMM's estimation of the next latent state (True value: $\pi_{i,t}$, Estimated value: $\hat{\pi}_{i,t}$) given the immediate past latent state ($\pi_{i,t-1}$) as: 
\begin{equation}
    \mathcal{L}_{\text{MSE}} = \sum_{t=\tau}^{T} \parallel \pi_{i,t} -  \hat{\pi}_{i,t} \parallel_{2}
\end{equation}

\item Categorical Cross Entropy loss over the probability distribution of both the predicted and the actual ($c_{i,t} \in \mathbb{R}^\rho$) condition via a dense layer with $\rho$ (number of conditions) as a hyper-parameter:
\begin{equation}            \mathcal{L}_{\text{Conditional}} = - \frac{1}{\rho}\sum_{t=\tau}^{T} \mathcal{P}_{\text{model}}(\pi_{i,t}) \log (\mathcal{P}_{\text{model}}(\hat{\pi}_{i,t}))
\end{equation}
\end{enumerate}
 where $\mathcal{P}_{\text{model}}()$ represents the Conditional Network consisting of a deep neural network such that $c_{i,t} = \mathcal{P}_{\text{model}}(\pi_{i,t})$. For details on the training complexity for training the DMM, refer to Appendix \ref{appendix:trainingflow}. The total transition loss of the TM is given by:
\begin{equation}
    \mathcal{L}_{\text{Transition}} = \mathcal{L}_{\text{MSE}} + \mathcal{L}_{\text{Conditional}}
\end{equation}
Where $\mathcal{L}_{\text{MSE}} $ characterizes the loss of the DMM. TM is pre-trained before training the complete AFN network. 
\subsubsection{Conditional Variational Auto-Encoder with Latent SOM (ConVAE - SOM)}
AFN uses a conditional VAE (cVAE) which leverages the conditions generated by the Transition Module. This ensures that the time-series data having the same Markov conditions is mapped to similar regularized latent space. SOM based topology is learnt over the entire conditional latent space as described in ~\cite{som-vae,DPSOM}. Input to the cVAE is $x_{i,t}~||~c_{i,t}$, where $c_{i,t}$ refers to the discrete condition for the $i^{th}$ player at time $t$ derived from the Transition Module, and $||$ denotes vector concatenation operator. cVAE encodes each input $x_{i,t}$ to a latent encoding $z_{e_{i,t}} \in \mathbb{R}^{m}$  ($m < d$). We will refer to $z_{e_{i,t}}$ as $z_{i,t}$ for brevity. The SOM training step then allocates each encoding $z_{i,t}$ to a cluster-centroid embedding $z_{q} \in \mathbb{R}^{m}$. Total number of SOM cluster embeddings is a hyper parameter and is derived empirically. In AFN, the SOM space is set to a $8 \times 8$, 2-D space, which results in 64 discrete SOM clusters. We use the same ELBO loss as in \cite{DPSOM} for cVAE, however while passing the $c$ condition to the decoder for the reconstruction we apply a gradient stopping on the TM predictions. This is being done to prevent SOM or cVAE network losses from influencing the conditions generated by the Transition Module. We borrow the loss formulation $\mathcal{L}_{\text{T-DPSOM}}$ from \cite{DPSOM}:
\begin{equation}
    \mathcal{L}_{\text{T-DPSOM}} = \beta \mathcal{L}_{\text{SOM}} + \gamma \mathcal{L}_{\text{Commit}} + \theta \mathcal{L}_{\text{Reconstruction}} + \kappa \mathcal{L}_{\text{Smoothness}}
\end{equation}
where $\beta, \gamma, \theta, \kappa$ are hyperparameters. To learn more about the settings of these various hyperparameters in our context as well as the process used to derive them, please refer to the Appendix Section \ref{appendix:parametertuning} 

\subsubsection{Intelligent Forecasting Module}
The Intelligent Forecasting Module (IFM) consists of:
\begin{enumerate}
    \item LSTM network with an Attention layer \textbf{(AL)} over the temporal latent encodings ($z_e$)
    \item Damping Factor Network \textbf{(DF)} which provides `leeway' for allocating distant SOM clusters, in case of a change in condition from the Transition Module.
    \item Forecasting Fine-Tuning \textbf{(FFT)} via MSE loss on the actual future predictions ($\hat{x}_{i,t+j}, j \in \lbrace 1, 2 \dots , h \rbrace$)
\end{enumerate}
LSTM based forecasting module generates the predictions for the next time steps within the latent space in form of the probability distribution over the next latent encoding, p($z_{i,t+1} \mid z_{i,t}$) using log-likelihood loss between actual and predicted $z$ distributions. We incorporate a temporal Attention layer on top of the LSTM in order to improve model performance and also be able to extract dominant time-steps that led to the forecast. To aid accurate forecasting while preserving a smooth trajectory, AFN introduces a Damping Factor, that is introduced to the LSTM prediction loss as: 
\begin{equation}
    \mathcal{L}_{\text{Pred}} = - \sum_{i=1}^{N} \sum_{t=1}^{T-1} \log p(z_{i,t+1} \big| z_{i,t}) ~\mathcal{D}_{\text{model}}(\pi_{i,t},\hat{\pi}_{i,t+1})
\end{equation} 
where $p(z_{i,t+1} \big| z_{i,t})$ refers to the probability distribution over the next latent encoding, given the current latent encoding. The Damping Factor Network, $\mathcal{D}_{\text{model}}()$ is tasked with a new objective to learn a damping factor between the latent state ($\pi_{i,t}$) and next latent state predicted by the DMM ($\hat{\pi}_{i,t+1}$) for $t \in \lbrace 1 \dots,~ T-1 \rbrace $.  To train the damping factor network, $\pi_{i,t}$ and $\hat{\pi}_{i,t+1}$ are passed through a fully connected neural network that yields a sigmoid-scaled output between $[0,1]$. In a case where there is a sudden data pattern shift, causing the transition to a far away SOM grid position in consecutive time steps, the low log-likelihood would generate a high $\mathcal{L}_{\text{Pred}}$ value (penalty). Hence, to compensate in favour of $\mathcal{L}_{\text{Pred}}$, AFN offers the LSTM an option to dampen this effect by simultaneously training the damping factor network. The damping loss is back-propagated to the SOM training layer, which also realigns its SOM cluster space to preserve interpretability. LSTM network generates next latent encoding, $z_{i, t+1} \sim p(z_{i,t+1} \big| z_{i,t})$ which is then passed to the decoder to generate $\hat{x}_{i,t+1}$. Forecasting fine tuning is done post this LSTM prediction step, in order to fine-tune the predictions in the actual feature space (unlike the $\mathcal{L}_{\text{Pred}}$ that works in the latent feature space). The Forecasting Fine-Tuning loss ($\mathcal{L}_{\text{Forecasting}}$) is defined as:
\begin{equation}
    \mathcal{L}_{\text{Forecasting}} = || x_{i,t+1} - \hat{x}_{i,t+1} ||_{2}
\end{equation}

\noindent The final loss function in AFN is given as:
\begin{equation}
    \mathcal{L}_{\text{AFN}} = \mathcal{L}_{\text{T-DPSOM}} + \tau \mathcal{L}_{\text{Transition}} + \eta \mathcal{L}_{\text{Pred}} +  \mathcal{L}_{\text{Forecasting}}
\end{equation}
where $\tau, \eta$ are again the hyperparameters.

\subsubsection{Addressing Challenges Related to SHaP over Temporal Data:}
\label{sec:shap_how}
Traditionally, SHAP based models identify feature weights for a prediction which is either a multi-class classification or a regression task. Our network is unsupervised. So, in AFN we let the SHAP model regress its Shapley values on the SOM cluster identities instead as the y-label and the actual game play features of the data point representing the corresponding SOM Cluster centroid as the x-value vector. We retain top \textit{5} Shapley values per SOM cluster (8x8 -64 in our case). But, this does not solve the two challenges related to capturing the temporal dependencies and steady focus on the most attentive feature across the entire forecasted period. 
For a player who is predicted to turn risky at a future time step t+f , t being the current time step and f $\le$ h, AFN publishes one or more attentive time steps for the forecast, via the \textit{AL}. AFN then uses the top 5 Shapley values at each of the attentive time steps (corresponding SOM centroids) and measures the extent of acceleration amongst each of the respective features. These features are then ordered on their average Shapley weights values followed by the magnitude of acceleration during the period of attention. The ranked list is published, from which we chose the topmost to start with.  


\section{Evaluation}
\subsection{Data}
For preparing our data, we picked up 200k players with a train-test split of 0.75. For each of the player, we extracted 90 days of continuous time series data across different times of the year. We fetch 19 features to represent their game-play style and indulgence. We could categorise these features into 3 dimensions: Time, Money, Desperation. Few of these are illustrated in Table\ref{tab:players_data_features}.

\begin{table}[h]
\centering
\small
\begin{tabular}{cl}
    \toprule
    \textbf{Dimension} & \textbf{Feature}  \\
    \midrule
    \multirow {2}{*}{Time} & Total time spent per day  \\
    & Count of late night games\\
    \midrule
    \multirow {3}{*}{Money} & Amount of cash added for playing  \\
    & Count of cash games played \\
    & Win percentage\\
    \midrule
    \multirow {3}{*}{Desperation} & Count of requests of increasing deposit limit \\
    & Invalid declaration of the win \\
    & Continue to play on bad cards (drop adherence) \tablefootnote{Drop adherence evaluates player's ability to judge if the initial rummy cards qualify as a bad hand\cite{kddgames,pakddgames}}\\
    \bottomrule
\end{tabular}
\caption{\small \textmd{Sample Game Play Features and their Broad Categorization}}
\label{tab:players_data_features}
\vspace{-25pt}
\end{table}


\subsection{Randomness in Dataset}
\label{sec:random_evaluation}
To validate our hypothesis of inherent high degree of  non-smoothness (randomness) in real world online player game-play data compared to frequently used open-source MTS datasets (such as ETTh), we perform the following statistical and graphical tests:

\subsubsection{Runs Test}
Runs test\cite{runs_test} is a non-parametric statistical test that checks randomness hypothesis for a two-valued data sequence. This test is based on the null hypothesis that the data is random. Table \ref{tab:runs_autocorr} shows p-values of the runs test obtained for each of the datasets. $n$ represents number of samples considered for each experiment. For players data, a sample represents a player, whereas for open source data it is a slice of the entire sequence, lengths of samples being the same in both the cases - 13.  We repeated the sampling process 1000 times. We report the mean and the standard deviation.

\begin{table}[!h]
    \centering
    \small
    \begin{tabular}{|l|l|l|l|l|l|l|}
    \hline
        Dataset & \multicolumn{3}{l|}{Runs Test} &  \multicolumn{3}{l|}{Auto Correlation Test} \\ \cline{2-7}
        ~ & n=30 & n=100 & n=300 & n=30 & n=100 & n=300 \\ \hline
        ETTh & 0.376$\pm$ & 0.353$\pm$ & 0.282$\pm$ & 0.246$\pm$ & 0.258$\pm$ & 0.267$\pm$ \\
        ~ & 0.109 & 0.192 & 0.125 & 0.111 & 0.128 & 0.122 \\ \hline
        ECL & 0.303$\pm$ & 0.296$\pm$ & 0.352$\pm$ & 0.245$\pm$ & 0.237$\pm$ & 0.234$\pm$ \\ 
        ~ & 0.188 & 0.174 & 0.197 & 0.069 & 0.088 & 0.085 \\ \hline
        WTH & 0.394$\pm$ & 0.300$\pm$ & 0.245$\pm$ & 0.286$\pm$ & 0.299$\pm$ & 0.318$\pm$ \\ 
        ~ & 0.208 & 0.256 & 0.121 & 0.111 & 0.103 & 0.100 \\ \hline
        \textbf{PD} & \textbf{0.439$\pm$} & \textbf{0.452$\pm$} & \textbf{0.453$\pm$} & \textbf{0.084$\pm$} & \textbf{0.077$\pm$} & \textbf{0.081$\pm$} \\ 
        ~ & \textbf{0.112} & \textbf{0.065} & \textbf{0.032} & \textbf{0.023} & \textbf{0.007} & \textbf{0.018} \\ \hline
        
    \end{tabular}
    \caption{\small \textmd{In runs test, Players data (PD) has maximum p-value across all the datasets describing a high degree of randomness compared to open-source datasets. Similarly, for Auto-correlation test, PD has the least correlation ratio implying a closeness towards a ``white-noise'' compared to the others.}}
\label{tab:runs_autocorr}
\vspace{-20pt}
\end{table}

        

\subsubsection{Auto-correlation Test}
Auto-correlation tests~\cite{forecast_book, auto-correlation} can be used to detect whether there is significant correlation between the observations and their lagged values. This provides insights into the presence of a structure or patterns in the time series data. In the presence of a trend in the data, short-term auto-correlations are typically positive and substantial, as nearby observations in time, share similar sizes. For seasonal data, auto-correlations are more prominent at seasonal lags (multiples of the seasonal frequency) compared to other lags. When data exhibit both trend and seasonality, a combination of these effects is observed. For a white noise series, we expect 95\% of the spikes in the ACF to lie within $\pm2/\sqrt{T}$ where $T$ is the length of the time series. Hence, lower ratios indicate higher resemblance to the white noise. Table \ref{tab:runs_autocorr} shows the ratio of number of lags that are above the threshold to the total lags. 

\subsubsection{Time-series Decomposition}
Time series decomposition~\cite{statsmodel, stl_decomp} is a well-known concept. In order to separate distinct patterns for more insightful analysis, we decompose the the time series to obtain trend, seasonal and residual components. We tried additive as well as multiplicative models to decompose. Figure \ref{fig:combine_decomp} in the Appendix Section \ref{appendix:time_series_decomposition} shows the decomposition of ETTh and Players data for variable (up to 150) timestamps. To quantify and compare the trend, seasonality and residual components across the datasets, we calculated proportion of variance explained by dividing the sum of squares of the individual components by the sum of squares of the original data. The value represents the percentage of the total variance explained by each component. Moreover, we also report the standard deviation of the residuals. This measures the overall level of noise and unexplained variation in the data. Table \ref{tab:seasonal_quant} compares these metrics on different datasets.
\begin{table}[!h]
    \centering
    \small
    \begin{tabular}{|c|c|c|c|c|}
    \hline
        Dataset & \multicolumn{3}{c|}{Explained Variance} & Residual \\ \cline{2-4}
        ~ & Trend & Seasonality & Residual & Standard Deviation \\ \hline
        ETTh & 91.684 & 2.167 & 1.653 & 0.079\\ \hline
        ECL & 87.927 & 3.371 & 3.743 & 0.088\\ \hline
        WTH & 93.564 & 0.131 & 0.297 & 0.036\\ \hline
        \textbf{PD} & \textbf{83.190} &\textbf{0.120} & \textbf{14.583} & \textbf{0.177}\\ \hline 
    \end{tabular}
    \caption{\small \textmd{Open-source datasets exhibits a higher proportion of explained variance from trend and seasonality compared to Player Data (PD). The residuals in PD exhibit a higher standard deviation than others, indicating a greater level of unexplained noise and volatility.}}
\label{tab:seasonal_quant}
\vspace{-20pt}
\end{table}


\subsection{Multi-Variate Time Series Forecasting}
Having established the inherent randomness in our players data, we check how several conventional as well as more sophisticated time series forecasting methods perform on this data. We input 7 weeks of data for each user and evaluate on 6 weeks of future predictions. This is done so as to ensure maximum Time to Survive (TTS, see Section \ref{sec:real_world_evaluation}) with minimum historical dependency. The results, summarized in Table \ref{tab:mse_exp}, show that AFN clearly outperforms the others even when the end goal is simply multi-variate, multi-timestep forecasting on our data. 

\begin{table}[h]
    \centering
    \begin{tabular}{lcccc}
    \toprule
        \textbf{Model} & \textbf{MAE} & \textbf{MSE} & \textbf{RMSE} & \textbf{MAPE}  \\ 
    \midrule
        Fast Fourier Transform & 0.17 & 0.06 & 0.22 & 1042.59  \\ 
        FB Prophet\cite{prophet} & 0.18 & 0.06 & 0.22 & 400.76  \\ 
        NLinear\cite{ndlinear} & 0.19 & 0.04 & 0.20 & 598.26  \\ 
        DLinear\cite{ndlinear} & 0.20 & 0.05 & 0.22 & 716.14  \\ 
        TCN\cite{temporal_convnet} & 0.20 & 0.04 & 0.21 & 694.15  \\ 
        \hdashline
        ARIMA & 0.26 & 0.12 & 0.29 & 508.31  \\ 
        XGBoost\cite{xgboost-nips} & 0.22 & 0.09 & 0.25 & 463.86  \\ 
        T-DPSOM\cite{DPSOM} & 0.16 & 0.04 & 0.21 & 401.88  \\ 
        \textbf{AFN} & \textbf{0.12} & \textbf{0.03} & \textbf{0.18} & \textbf{355.16} \\ 
    \bottomrule
    \end{tabular}
    \caption{\small \textmd{Evaluation of MTS models on players data. The models above the dotted line provide no actionable interpretations while the models below offer varying degrees of interpretation and explainability. SOM-VAE~\cite{som-vae} does not offer forecasting feature.}}
    \label{tab:mse_exp}
    \vspace{-20pt}
\end{table}

\begin{figure}[b]
  \includegraphics[width=\linewidth]{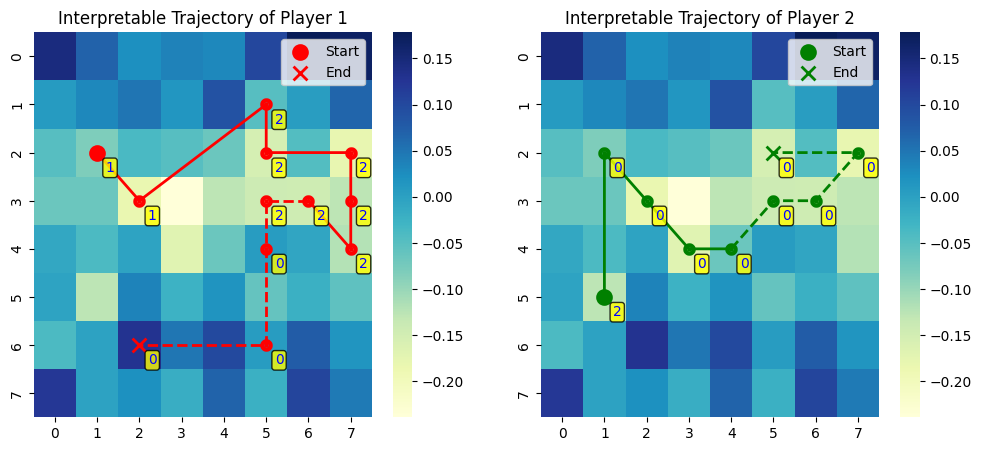}
  \caption{\small \textmd{Trajectory of Player 1 and Player 2 on a 2D SOM grid of risk-related clusters. AFN forecasts the trajectory of Player 1 towards the darker(risky) clusters, while it forecast a smooth trajectory on the lighter(healthy) clusters for Player 2. These maps enables the interpretability by visualising the state of player at each time step.}}
\label{fig:interpret_trajec}
\end{figure}

\subsection{Interpretability and Explainability}
Now we illustrate how the individual components of AFN contributes to the overall interpretability and explainability with user-specific trajectories as our working examples.

\begin{figure}[b]
  \includegraphics[width=\linewidth]{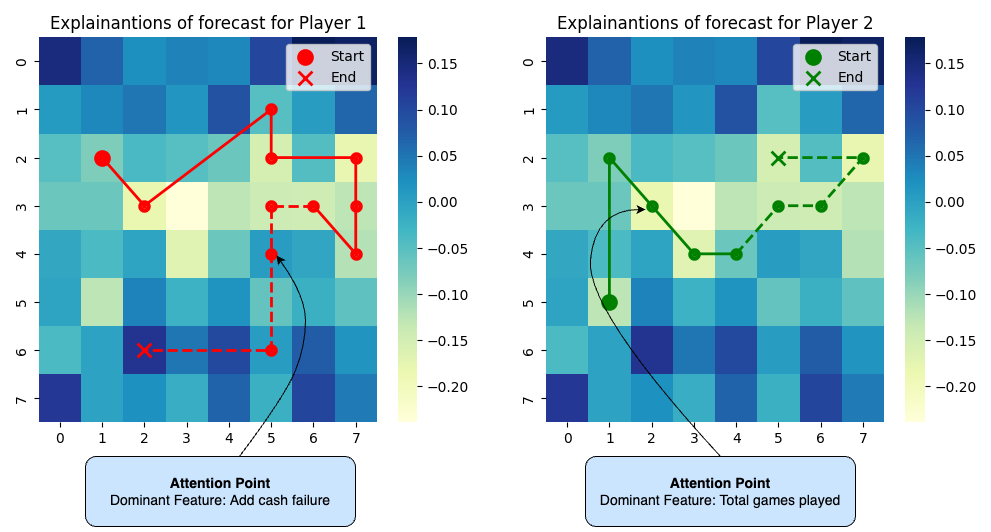}
  \caption{\small \textmd{Explanations of final forecast for Player 1 and Player 2 by showing the respective Attention Point and Dominant Feature respectively.}}
\label{fig:exp_trajec}
\end{figure}

\subsubsection{Interpretability:}
For establishing interpretability, we sample 500 users from test set. We used stratified random sampling by creating the buckets based on their last 60 days engagement level. These players were then randomly sampled from each stratum to ensure representation from all engagement categories. AFN forecasted their onward trajectory for the next 42 days. We filtered out the players who were in the same/neighbouring clusters for at least 70\% of their time (indicating same paced and healthy game play over time). Out of the remaining 68 players, we picked 2 random players that had a progressive journey into different SOM clusters. Its important to note that this could be due to movement to other game play patterns and not necessarily indicate an overindulgence. The interpretability of their journey is demonstrated by: 

\textbf{i. Visualisation of trajectory on SOM cluster grid:}
SOM embeddings generate SOM centroids over the ConVAE-SOM latent encodings. These embeddings posses topological relationship and can be visualized as a 2-D grid. Figure \ref{fig:interpret_trajec} shows a heatmap on a 8x8 grid with 64 SOM clusters. Each cluster is more similar to its neighbourhood clusters than others. This grid provides a foundation to visualise any player's trajectory. The color of this heatmap signifies indulgence level (darker being more indulgent). Level of risk or indulgence is generated using ScarceGAN ~\cite{scarcegan}. Figure \ref{fig:interpret_trajec} we've visualised player's known trajectory with solid lines and forecasted one with dotted lines. The red trajectory indicates a forecast that the player will move towards over-indulgence. The other player (green trajectory) however starts in a similar fashion initially but continues with constant momentum and hence is seen to move smoothly with all centroids showing low heat (lighter colours) and is declared to be playing with healthy habits for the future 42 days. 

\textbf{ii. Conditions derived from Transition Module:}
Transition module generates a condition value which represents the data pattern at the specific timestamp. Firstly, a change in the condition can be used to interpret randomness in the data. Secondly as seen in the Figure \ref{fig:interpret_trajec}, the red trajectory is associated with varying conditions and also specially less of condition 0. Whereas the the green trajectory sticks to a single condition most of the times and see more of condition 0. We further quantify the TM based condition driven interpretability in the Appendix in the Section \ref{appendix_quantifyInterpretability}. 

\subsubsection{Explainability:}
In order to provide reasoning of their forecast, we clubbed Shapley values and self-attention as discussed in Section \ref{sec:shap_how}. 

\textbf{i. Identifying significant events/time steps:}
Attention weights are used to identify time steps where the model attributed its attention for forecasting the future time steps. In fig \ref{fig:exp_trajec}, the forecasted journey is into the darker region which signifies over-indulgence. The attention point on $10^{th}$ time step implies that for this user $10^{th}$ time step majorly contributed in predicting overindulgence. In a proactive decision-making scenario, these attention points offer opportunities for an intervention. Figure \ref{fig:exp_trajec} also shows that the second player's healthy onward trajectory is primarily attributed to the player reducing the number of games played per day on the platform. This \textit{self-moderation} by the player was detected by AFN at the $3^{rd}$ (past) time step.  

\textbf{ii. Extracting the responsible features for forecast:}
Table \ref{tab:variable_imp} shows the top 5 features for 4 different SOM clusters (randomly chosen). 

\begin{table}[!ht]
    \centering
    \small
    \begin{tabular}{|lll|ll|}
        \hline
         \textbf{SOM Cluster (1,2)} & ~ & ~ &  \textbf{SOM Cluster (3,5)} & ~ \\ 
         \hline
        Add cash failure & +1.118 & ~ & Change in daily limit & +1.006 \\ 
        Win Ratio & -1.058 & ~ & Add cash transaction & -0.907 \\ 
        Drop adherence & +1.009 & ~ & Modes of payment & +0.784 \\ 
        Invalid declaration & -0.830 & ~ & Net winnings & -0.550 \\ 
        Change in daily limit & -0.578 & ~ & Total games played & +0.486 \\
        \hline
    \end{tabular}
    \caption{\small \textmd{Sample summary of the Shapley values for two SOM clusters. Add cash failure, Win Ratio, and Drop Adherence emerge as the most influential factors, in cluster (1,2)}}
    \label{tab:variable_imp}
    \vspace{-25pt}
\end{table}

\begin{figure}[htbp!]
    \centering
      \begin{subfigure}{0.22\textwidth}
        \includegraphics[width=\textwidth]{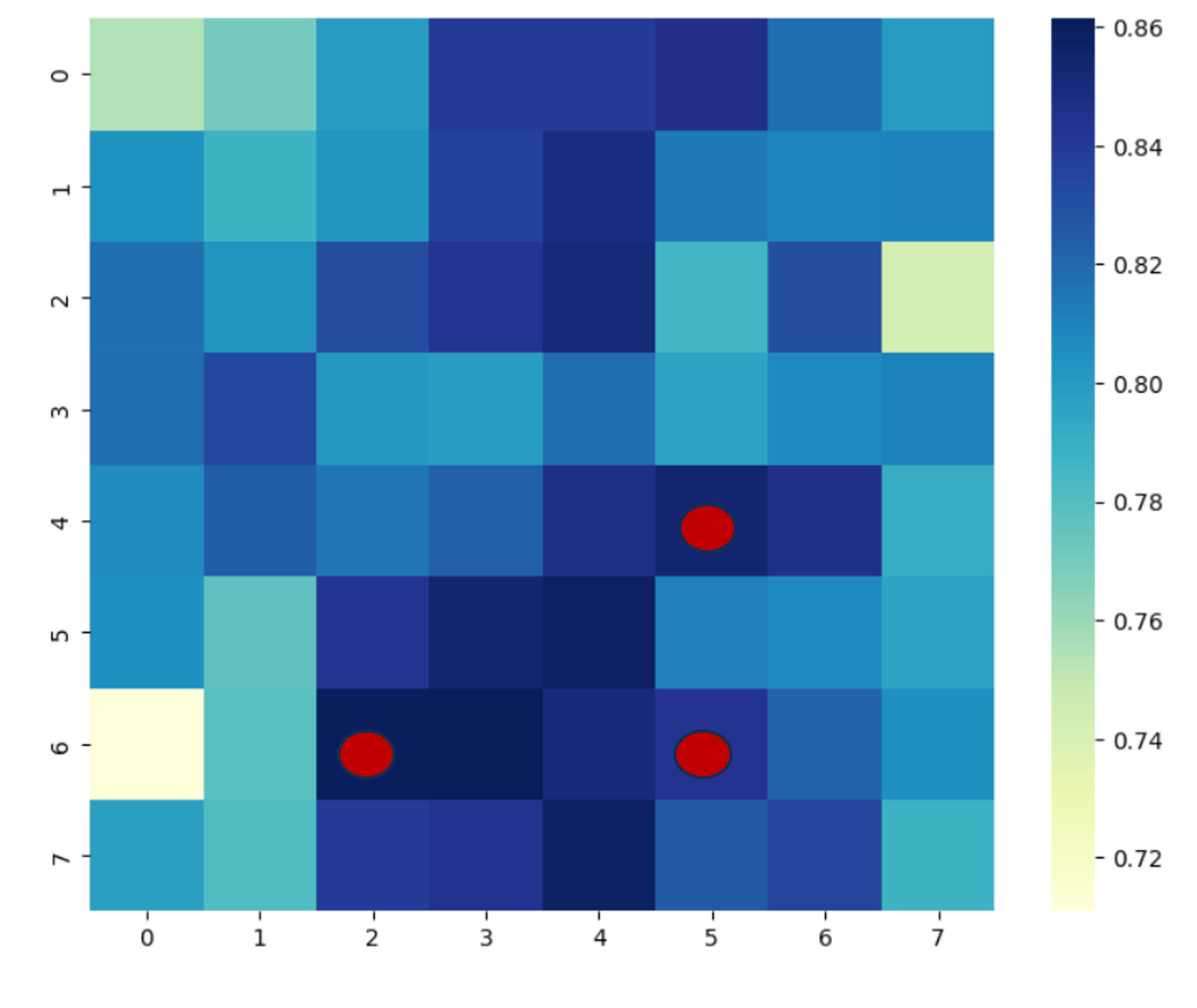}
          \caption{Dominant Feature}
          \label{SHAP_ordering-a}
      \end{subfigure}
      \hfill
      \begin{subfigure}{0.22\textwidth}
        \includegraphics[width=\textwidth]{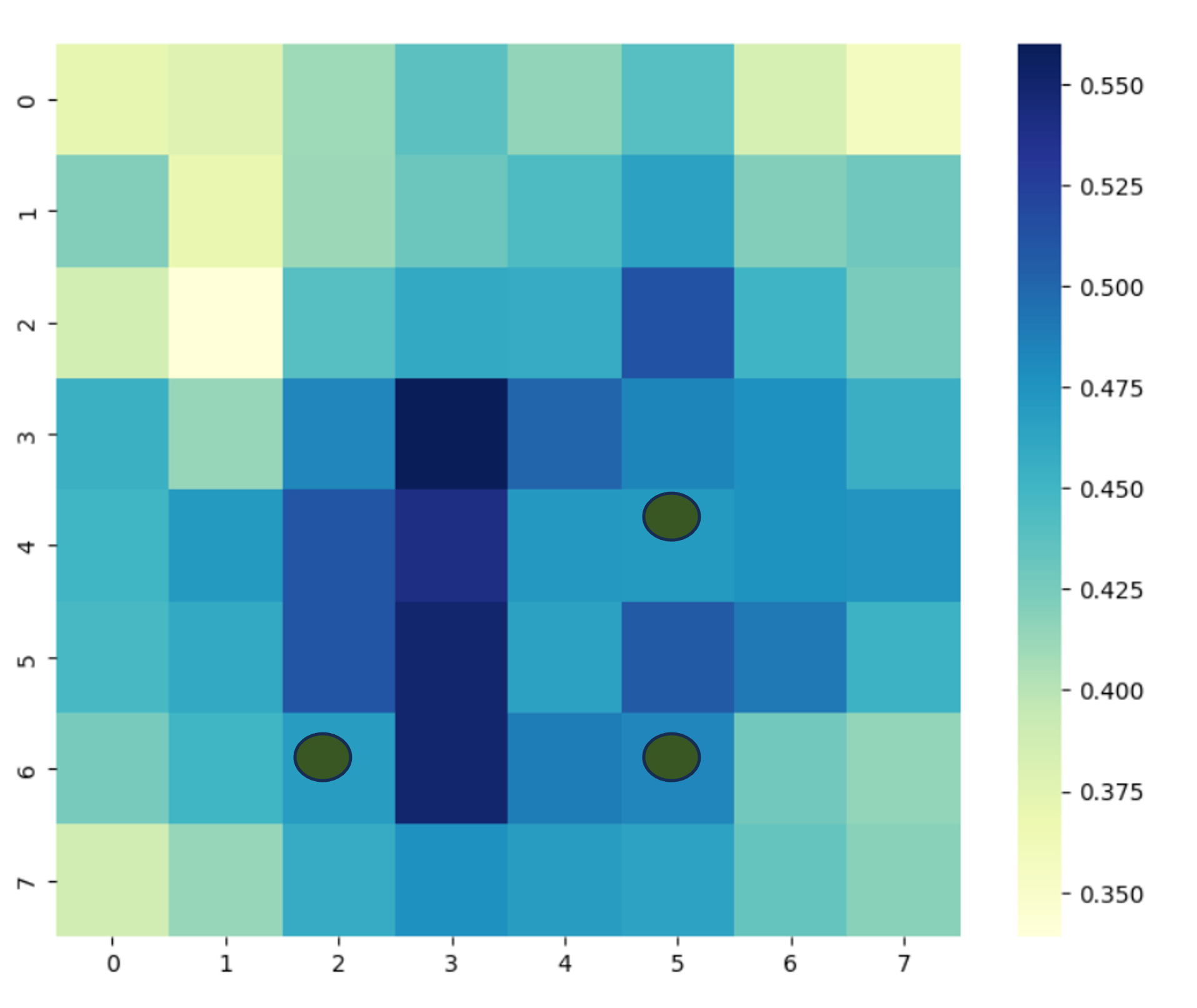}
          \caption{Second Dominant Feature}
          \label{SHAP_ordering-b}
      \end{subfigure}
\caption{\small \textmd{heatmap of Dominant and a Non-Dominant Feature. Dominant feature clearly shows higher Shapley magnitude throughout the attentive period for the player demonstrated in Figure \ref{fig:exp_trajec}}}
\label{SHAP_ordering}%
\vspace{-20pt}
\end{figure}

In this paper, we illustrate results on the first important feature obtained through the ranking process discussed in the Section \ref{sec:shap_how}.  Figure \ref{SHAP_ordering} refers to the two top attentive features for the player in the red trajectory in the Figure \ref{fig:exp_trajec}. Figure \ref{SHAP_ordering-a} refers to the heatmap of the most attentive feature. We see that the Shapley value of this feature is constant high and shows better acceleration (red dots) compared to ones in the heatmap of the second dominant feature (green dots), win ratio. The scale of the heatmap in both the graphs signify their Shapley values for each of the SOM centroids. The Dot locations corresponds to the attention points in the SR (red) trajectory in the Figure \ref{fig:exp_trajec}. We understand that, though, the heatmaps of each feature remains fixed after the AFN training, based on the trajectory, dominant attentive feature of a players' could vary and hence it is personalized.

\subsection{Real-world evaluation}
\label{sec:real_world_evaluation}
In this section, we'll demonstrate the applicability of this entire framework for our gaming players. AFN intends to facilitate responsible gaming on our platform. Early detection of risky gaming pattern (over-indulgence, chasing losses etc.) can enable timely intervention and personalised support for players at risk. Based on the AFN forecast, each future timestamp is mapped to a corresponding SOM cluster. Leveraging these forecasted clusters, we transform the objective into a binary classification problem of \textit{Sustained Risky (SR)} vs \textit{Sustained Healthy (SH)} class. Given the dynamic nature of the platform, its highly likely to see highs and lows in risk scores in the subsequent time steps for any player. For each of the cluster, we generate a normalised risk score using ScarceGAN\cite{scarcegan} which represents the color scale of the heatmap. All SOM clusters that have values > 0.0 are identified as risky/dark clusters and remaining are identified as healthy/light clusters. For player to be identified as SR, we introduce \textit{Burst logic} where we define burst as consecutive allotment of dark clusters. A player is classified as SR only if there's a burst of size greater than threshold. We took threshold value as 2 for the Burst logic. All other possibilities are classified into SH class. At a user level, we define \textit{Time to Survive (TTS)} as the time difference between burst start time step and the present time step (T). TTS signifies the minimum amount of time left to take preventive measures and restrict user from becoming Sustained Risky (SR). We also define \textit{Verbosity} as proportion of users classified as SR.

\subsubsection{Evaluating Risk Mitigation through AFN}
We contend that, for players predicted to be SR on our platform, proactive and timely intervention along the right aspect of their game play could actually help prevent overindulge. Table \ref{tab:verbos_exp} shows an example for a cohort of 60 players, forecasted to be SR. Magnitude of the respective attentive feature for each player was reduced synthetically in small steps. Reduction was only done at the attentive time step. At each reduction we infer player's onward journey using AFN. We observe an evident reduction in overindulgence volume along with an increase in the TTS percentage.  
\vspace{-5pt}
\begin{table}[!ht]
\small
\begin{tabular}{ccc}
\toprule
 \textbf{Reduction Factor} &  \textbf{Change in SR Volume} &  \textbf{Change in TTS} \\
\midrule
    5 \%  & - 1.37 \%  & 1.39 \% \\
    10 \% & - 2.75 \%  & 1.28 \% \\
    20 \% & - 8.96 \%  & 1.90 \% \\
    30 \% & - 9.65 \%  & 2.68 \% \\
    40 \% & - 13.10 \% & 3.64 \% \\
    50 \% & - 16.55 \% & 5.03 \% \\
    60 \% & - 17.93 \% & 5.33 \% \\
    70 \% & - 18.62 \% & 6.22 \% \\
\bottomrule
\end{tabular}
\caption{\small \textmd{Change in Verbosity and Time to Survive (TTS) on reducing most dominant feature of attention at the right time}}
\label{tab:verbos_exp}
\vspace{-15pt}
\end{table}

\subsubsection{Risk mitigation via personalized intervention}
\begin{figure}[!h]
\vspace{-5pt}
    \includegraphics[width=0.5\linewidth]{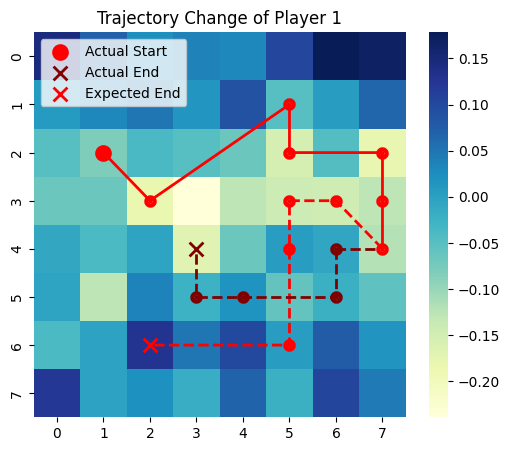}
  \caption{\small \textmd{Illustration of how a forecasted SR trajectory for a player can be restored to a SH via timely and personalized intervention}}
  \label{fig:trajectory_change_kdd_p1}
\end{figure}
We pick an example of personalized intervention from the cohort in Table \ref{tab:verbos_exp}, shown in Figure \ref{fig:trajectory_change_kdd_p1}. Player was predicted to be SR at time step 7, with next 5 time steps in red dotted lines. This player was intervened at the $10^{th}$ time step, where the attention was drawn by AFN along the responsible attentive feature \textit{Add cash failure} (acf). We applied 20\% reduction in the player's \textit{wallet} add cash limit. Figure \ref{fig:trajectory_change_kdd_p1} also shows the entire ``actual'' trajectory of the player with red solid (7 steps) and brown dashed line (5 steps). This is an example of how overindulgence can be prevented with intervention on the right feature (cash limit, in this case), at the right time step.

\subsubsection{Evaluating our classification model}
We also evaluate our AFN classification framework with ScarceGAN\cite{scarcegan}, retrospectively. 
We learn from Table \ref{tab:pr_comp_with_scarcegan} that out the total volume predicted risky by ScarceGAN on a given date, AFN identifies over 23\% of those in advance with and average TTS of 4 weeks (30 days) and is 43\% precise in its predictions. AFN outperforms T-DPSOM based SOTA model by 100\%.
\begin{table}[h]
\small
\begin{tabular}{cccc}
\toprule
  \textbf{Model} & \textbf{Precision} & \textbf{Recall} & \textbf{Verbosity} \\ 
\midrule
  T-DPSOM~\cite{DPSOM} & 0.2416 & 0.1113 & 0.4563 \\
  AFN & \textbf{0.4306} & \textbf{0.2383} & \textbf{0.3377} \\
\bottomrule
\end{tabular}
\caption{\small \textmd{In-Production Performance of AFN for Early Risk Prediction}}
\label{tab:pr_comp_with_scarcegan}
\vspace{-25pt}
\end{table}

\subsection{Ablation Study}
In our approach, there are four novel components: Transition - \textbf{TM}, Attention Layer - \textbf{AL}, Damping Factor Network- \textbf{DF} and Forecasting Fine-tuning - \textbf{FFT}. In the Table \ref{tab:mse_ablation_results}, AFN refers to \textbf{TM + AL + DF + FFT} i.e. no ablation. We've performed ablation over these components to study their effect on the MSE and classification performance. 
Table \ref{tab:mse_ablation_results} shows the MSE results for each experiment (detailed graphs in Appendix \ref{appendix:ablation}).
\begin{table}[h]
\centering
\small
\begin{tabular}{ccccc}
\toprule
\textbf{Training Steps} & \textbf{Without} & \textbf{Without} & \textbf{Without} & \textbf{AFN} \\
~ & \textbf{TM} & \textbf{AL} & \textbf{DF} & ~\\
\midrule
MSE without FFT & 0.111 & 0.059 & 0.068 & \textbf{0.051}\\
\textbf{MSE with FFT} & 0.147 & 0.080 & 0.071 & \textbf{0.039}\\
\bottomrule
\end{tabular}
\caption{\small \textmd{MSE results on component ablation using Players data.}} 
\label{tab:mse_ablation_results}
\vspace{-25pt}
\end{table}

\section{Conclusion}
We have proposed AFN, a novel deep neural network for interpretable forecasts and explanations over the multi-dimensional and non-smooth temporal feature space. We use specialized components to create such forecasts for the chaotic, multi-source time series often encountered in business: (1) Transition Module consisting of DMM and Conditional Network addresses the randomness in the data by changing the conditions in case of a change in data pattern, (2) ConVAE-SOM network generates interpretable intermediate trajectories for the forecasts, and finally (3) Intelligent Forecasting Module that performs the actual forecasting by intelligently utilizing the Damping Factor Network to incorporate data pattern shifts, while maintaining explainability over the forecasts using the Attention Module and SHAP over the SOM clusters. 

We have incorporated AFN on our online gaming platform to proactively flag, would-be risky players. We provide personalized time and attentive feature based interventions to the players to prevent the eventualities. We have published our data and source code at ~\cite{AFNgit}. We contend that AFN is generally applicable for actionable forecasting in a variety of business domains and use cases.

There are two systematic exploration possibilities that one could extend forward from this work. Firstly, one could revisit the logic for reconstruction of a sample from the VAE based network to further improve MSE's. Secondly, one could explore a different latent space structure (apart from SOM), for example a 2D grid, by learning the data relationships as a graph and enforcing those learnings on the latent space.
\bibliographystyle{plain}

\newpage
\section{Appendix}
\begin{appendices}

\section{Implementation Details}

\subsection{Model Training Details}
\label{appendix:parametertuning}
For running various experiments on players data as well as other real world datasets, we procured AWS g4dn.xlarge (1x NVIDIA T4 GPU, 4 vCPUs, 16GB RAM) instance for $K=5$, $\rho=3$ and batch size of 128. The models parameters are described in Table~\ref{tab:model_config}. It took 1.7 hours to train the model on 150k train set users.

\begin{table}[h!]
    \centering
    \begin{tabular}{ll}
    \toprule
    \textbf{Parameter} & \textbf{Value} \\
    \midrule
    DMM & 2 linear layers: (500, 128) \\
    Condition Network & 1 linear layer: (16,) \\
    ConVAE-SOM Encoder & 3 stacked LSTM: (500, 500, 2000) \\
    ConVAE-SOM Decoder & 3 stacked LSTM: (2000, 500, 500) \\
    Damp Factor Network & 2 linear layers: (100, 10) \\
    LSTM Forecasting Module & LSTM: (100, ) \\
    with attention & \\
    \bottomrule
    \end{tabular}
    \caption{\textmd{Model Parameters Configuration}} 
    \label{tab:model_config}
    \vspace{-20pt}
\end{table}



When compared to SOTA transformer models, AFN exhibits a substantial reduction in the number of parameters required for effective forecasting. While transformers are known for their impressive performance across various domains, they tend to be parameter-intensive, often requiring massive computational resources. In contrast, AFN demonstrates a clever combination of Deep Markov Model and Damp Factor, allowing it to achieve comparable Mean Squared Error (MSE) results with a significantly smaller parameter count.

\subsection{Hyper-Parameter Selection}
\label{appendix:hyperparametertuning}
Hyper-parameters for AFN were chosen using the heuristics provided in \cite{DPSOM} and further tuned using Optuna\cite{optuna} open-source Python library. The following settings of the various weights were indicated to be optimal.
\begin{table}[h!]
\centering
\begin{tabular}{cc}
\toprule
\textbf{Hyper-Parameter} & \textbf{Value} \\
\midrule
Alpha ($\alpha$) & 10.0 \\
Beta ($\beta$) & 0.1 \\
Gamma ($\gamma$) & 0.3 \\
Kappa ($\kappa$) & 1.0 \\
Theta ($\theta$) & 0.1 \\
Eta ($\eta$) & 10.0 \\
Tau ($\tau$) & 75 \\
\bottomrule
\end{tabular}
\caption{\textmd{Final Hyper-Parameters used in Training}} 
\label{tab:evaluation_1}
\vspace{-20pt}
\end{table}

\subsection{Details of Ablation Experiments on AFN}
\label{appendix:ablation}
\begin{figure}[]
        \centering
        \begin{subfigure}[b]{0.5\textwidth}
            \centering
            \includegraphics[width=0.7\textwidth]{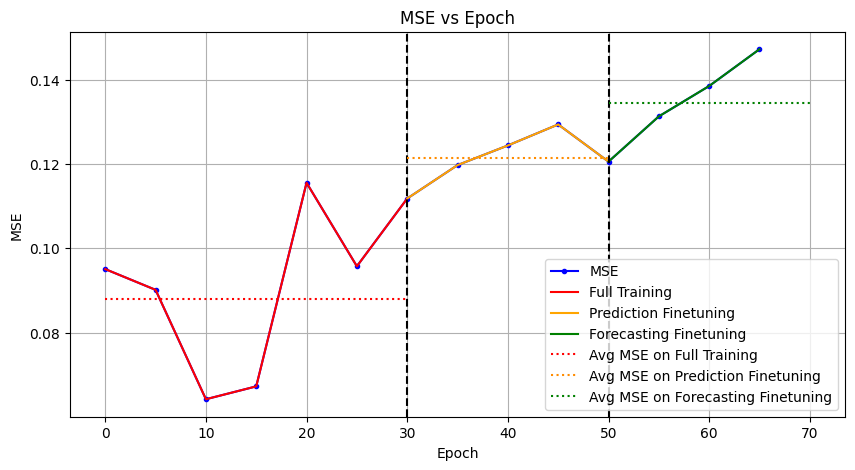}
            \caption{Without TM}
            \label{fig:MSE_linegraph_no_dmm4}
        \end{subfigure}
        \begin{subfigure}[b]{0.5\textwidth}  
            \centering 
            \includegraphics[width=0.7\textwidth]{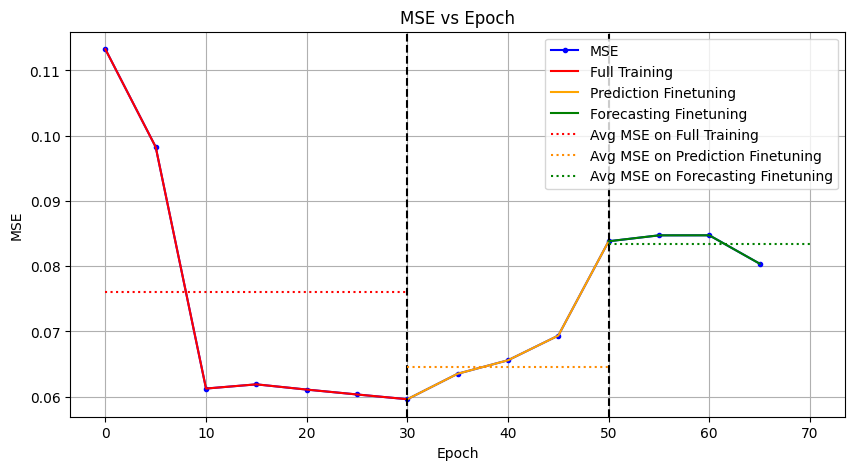}
            \caption{Without AL}
            \label{fig:MSE_linegraph_no_attention}
        \end{subfigure}
        \begin{subfigure}[b]{0.5\textwidth}   
            \centering 
            \includegraphics[width=0.7\textwidth]{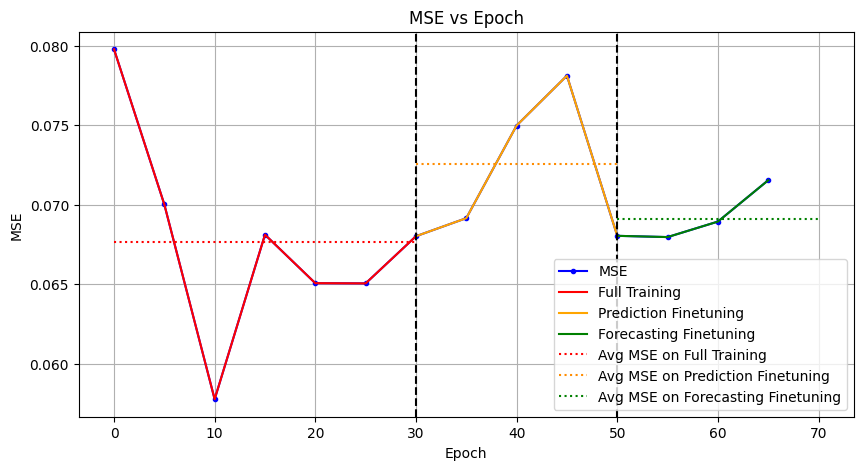}
            \caption{Without DF}
            \label{fig:MSE_linegraph_no_dampfactor}
        \end{subfigure}
        \begin{subfigure}[b]{0.5\textwidth}   
            \centering 
            \includegraphics[width=0.7\textwidth]{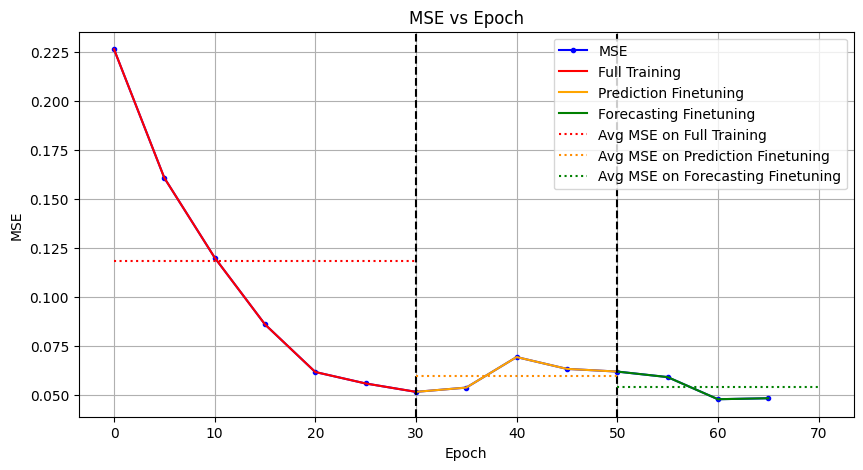}
            \caption{AFN}
            \label{fig:mean and std of net44}
        \end{subfigure}
        \caption{\textmd{MSE vs Epoch line-graphs on different ablations experiments.}} 
        \label{fig:mse_linegraphs}
\end{figure}

Figure~\ref{fig:mse_linegraphs} shows the variation of MSE in training workflow across all the ablation experiments. TM helps to capture the parameter distribution well, especially for non-linear player data and in its absence MSE could be high due lack of guiding loss function for non-linear data. Without AL, the MSE starts well, as TM is mapping the space well, however the rise in MSE during prediction fine tuning is significant due to lack of long term pattern purview.  Without DF, the model is not able to predict farther nodes  leading to higher MSE. AFN outperforms all the ablations.

\section{Training Complexity}
\label{appendix:trainingflow}
The time complexity of the training algorithm is discussed in Table~\ref{tab:time_complexity}, deriving inspiration from \cite{transformer}. 
\begin{table}[]
\centering
\begin{tabular}{cc}
\toprule
\textbf{Module Type} & \textbf{Complexity per Layer} \\
\midrule
Transition Module & $\mathcal{O}(k.d^2)$ \\
ConVAE-SOM & $\mathcal{O}(k.d^2)$ \\
Intelligent Forecasting Module & $\mathcal{O}(n.d^2)$ \\
\bottomrule
\end{tabular}
\caption{\textmd{Time complexity of AFN modules. $n=$ Sequence Length, $d=$ Representation dimension, $k=$ Number of hidden layers}} 
\label{tab:time_complexity}
\end{table}

\section{Extra Experimental Results}
\subsection{Comparison with MTS Forecasting SOTA}
Table~\ref{tab:sota_results} compares the forecasting performance of AFN with the SOTA transformer-based MTS forecasting models such as Crossformer~\cite{crossformer} and PatchTST~\cite{patch-tst}, on players data. We choose to compare with only these models since they have shown to have out-performed InFormer~\cite{informer}, AutoFormer~\cite{autoformer}, PyraFormer~\cite{pyraformer} and FEDFormer~\cite{fedformer}. Our findings indicate that while the AFN's MAE and MSE may not always be the highest, it offers comparable results with the added advantages of interpretability and explainability. For players data, PatchTST outperforms CrossFormer. 
\label{appendix:sota_results}
\begin{table}[]
    \centering
    \begin{tabular}{ccc}
    \toprule
        \textbf{Model} & \textbf{MAE} & \textbf{MSE} \\
    \midrule
        AFN & 0.124 & 0.039 \\
        CrossFormer & 0.067 & 0.010 \\
        PatchTST & 0.059 & 0.009 \\
    \bottomrule
        \end{tabular}
    \caption{\textmd{Comparison of AFN with SOTA models}} 
    \label{tab:sota_results}
    \vspace{-20pt}
\end{table}

\subsection{Auto-correlation tests}
\label{appendix:randomness}
\begin{figure}[]
        \centering
        \begin{subfigure}{0.225\textwidth}
            \centering
            \includegraphics[width=0.9\textwidth]{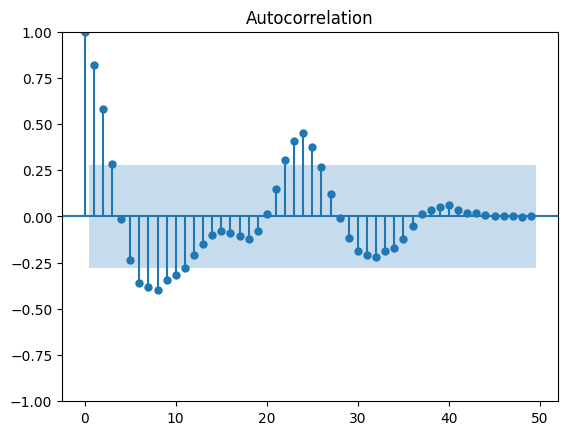}
            \caption[]%
            {\small ECL Dataset}
            \label{fig:correlogram_ecl}
        \end{subfigure}
        \hfill
        \begin{subfigure}{0.225\textwidth}  
            \centering 
            \includegraphics[width=0.9\textwidth]{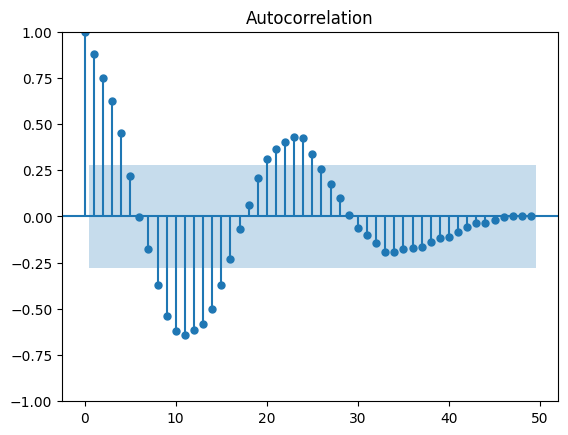}
            \caption[]%
            {{\small ETTh1 Dataset}}    
            \label{fig:correlogram_etth}
        \end{subfigure}
        \vskip\baselineskip
        \begin{subfigure}{0.225\textwidth}   
            \centering 
            \includegraphics[width=0.9\textwidth]{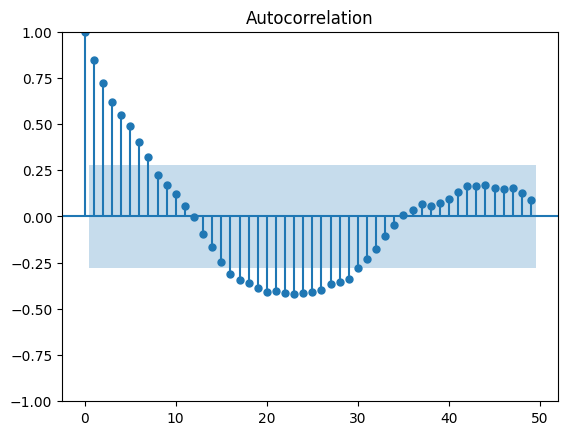}
            \caption[]%
            {{\small WTH Dataset}}    
            \label{fig:correlogram_wth}
        \end{subfigure}
        \hfill
        \begin{subfigure}{0.225\textwidth}   
            \centering 
            \includegraphics[width=0.9\textwidth]{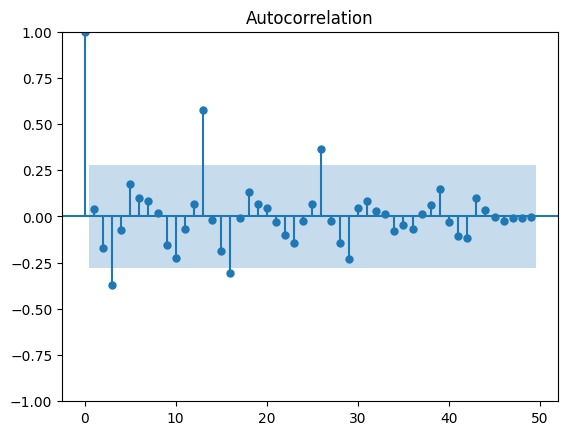}
            \caption[]%
            {{\small Players Data}}    
            \label{fig:correlogram_pd}
        \end{subfigure}
        \caption[]
    {\textmd{Auto-correlation Function (ACF) correlogram of different datasets}} 
        \label{fig:correlogram}
        \vspace{-25pt}
    \end{figure}

We compare correlograms in Figure~\ref{fig:correlogram} for players data with several open-source MTS datasets such as ECL, ETTh1 and WCL. We can clearly see that the lagged auto-correlations of the open-sourced datasets have higher instances and magnitude of significance.

\subsection{Visualization of Time Series Decomposition}
\label{appendix:time_series_decomposition}
\begin{figure}[]
  \includegraphics[width=\linewidth]{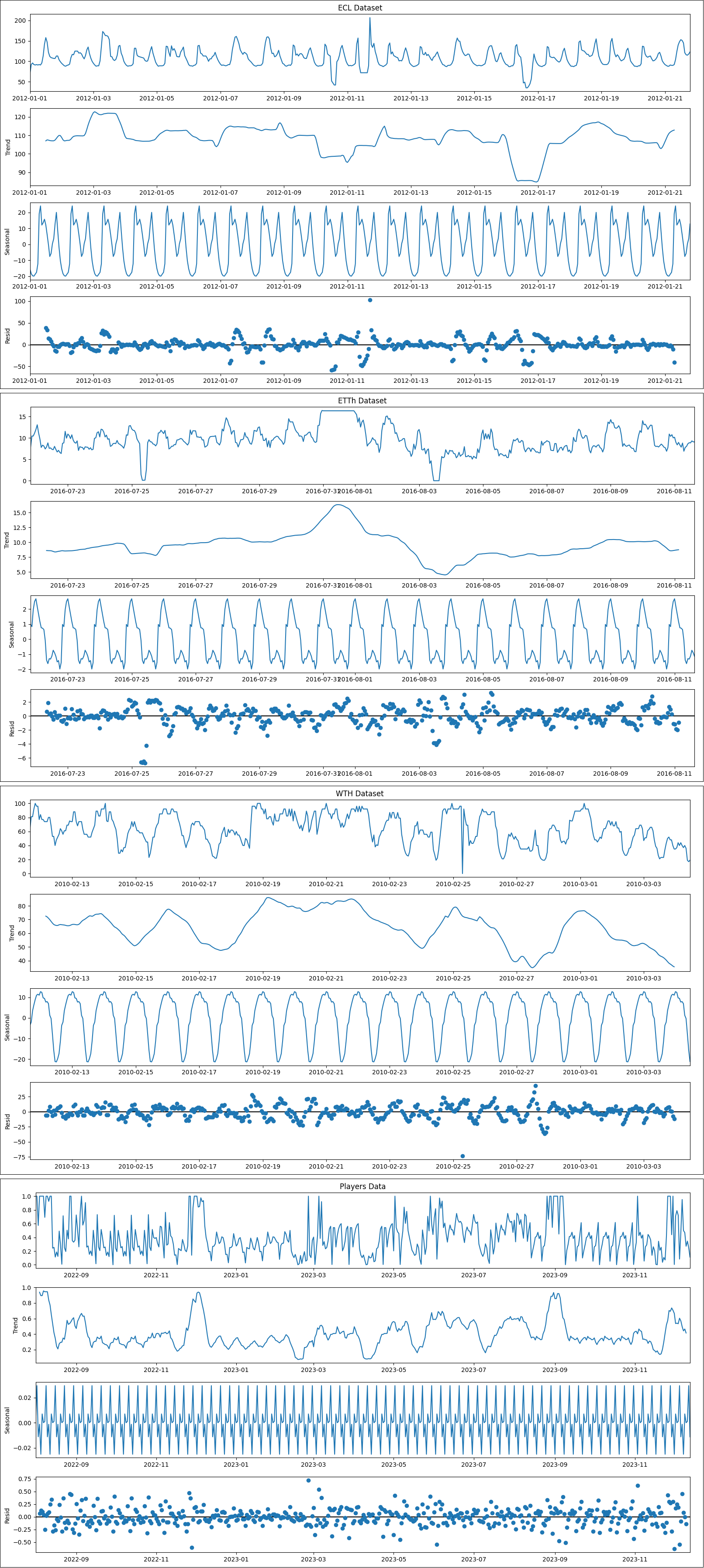}
  \caption{\textmd{These decompositions demonstrate the diverse characteristics of time series data. ECL, ETTh and WTH exhibits clear long-term trends and predictable seasonal patterns, making it amenable to forecasting and analysis. On the other hand, Players data lacks such structure, suggesting random fluctuations rather than predictable cycles. Furthermore, the residuals are more prominent (high deviation from the zero mean) in Players data indicating a high degree of randomness.  
}}
\label{fig:combine_decomp}
\end{figure}

Figure~\ref{fig:combine_decomp} describes the Time Series Decomposition of different datasets as mentioned in Section 4.2.3. In order to effectively compare the long term trend and seasonal patterns, we took 500 data points for each of these datasets. 

\subsection{Quantifying Interpretability and Visualisation of the SOM grid}
\label{appendix_quantifyInterpretability}
\begin{figure}[]
    \includegraphics[width=0.45\linewidth]{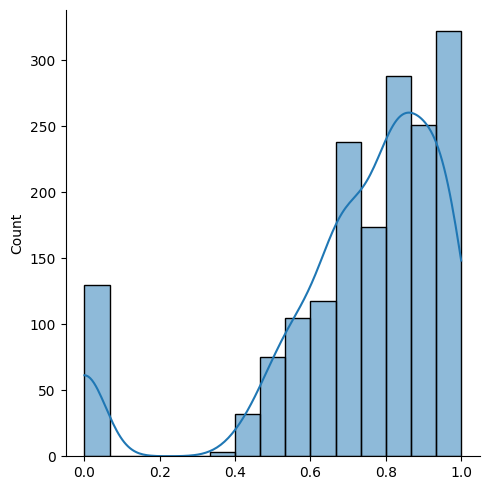}
    \caption{\textmd{Distribution of Pearson's Correlation values between non-linear trajectory and Markov condition change}}
    \label{fig:pearson-corr}
\end{figure}

\begin{figure}[]
    \includegraphics[width=0.38\textwidth]{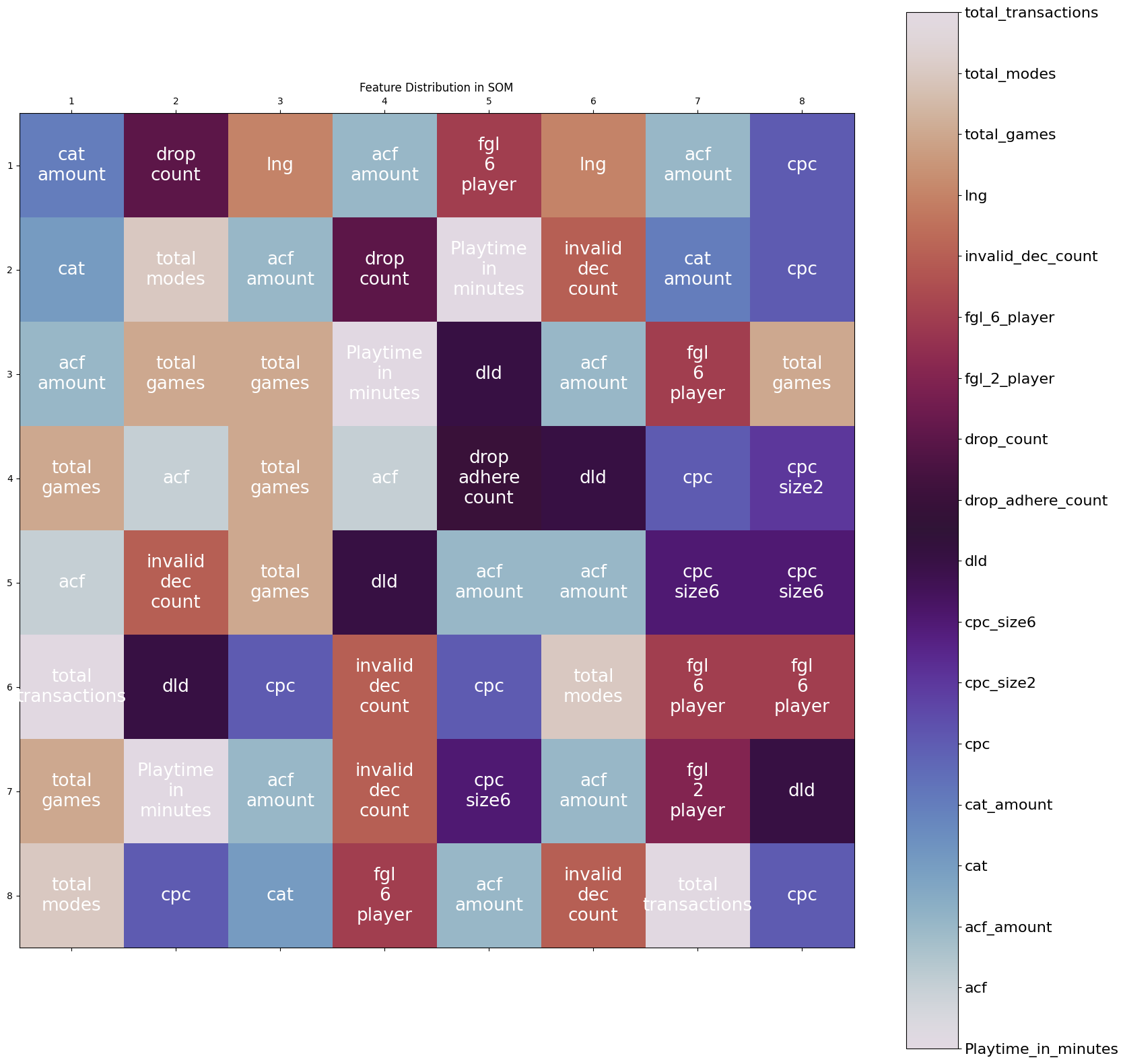}
    \caption{\textmd{Distribution of most important features across the SOM grid}}
    \label{fig:features_distribution_heatmap}
\end{figure}

With player data neighbouring transitions could be non-smooth, hampering interpretability. AFN leverages condition switch as an explanation to justify it. To validate if correlation between a long jump and condition switch is dependable, we created new time series by calculating shortest distance between the clusters assigned to the two successive time steps and correlated with the change in conditions predicted at each point. We found the Pearson's correlation~\cite{pearsons} to be primarily high and positive in most cases. Figure \ref{fig:pearson-corr} shows distribution of the correlation for about 3000 player samples that we forecasted. We take absolute value of the correlation, as both the positive and negative values contribute equally to the interpretations. The distribution is mostly centered towards a higher correlation coefficient with mean of 0.726 and a median value of 0.797. With mean and the median quite closer distribution seems less skewed to the left. This exercise assures that the transition to farther nodes are mostly associated with a change in the underlying Markov condition and hence is well interpretable. Figure~\ref{fig:features_distribution_heatmap} shows the most important features for each SOM grid. 


\end{appendices}

\end{document}